\documentclass[conference]{IEEEtran}
\IEEEoverridecommandlockouts
\usepackage{cite}
\usepackage{amsmath,amssymb,amsfonts}
\usepackage{algorithmic}
\usepackage{graphicx}
\usepackage{textcomp}
\usepackage{xcolor}
\def\BibTeX{{\rm B\kern-.05em{\sc i\kern-.025em b}\kern-.08em
    T\kern-.1667em\lower.7ex\hbox{E}\kern-.125emX}}

\usepackage{times}
\usepackage{soul}
\usepackage{url}
\usepackage[hidelinks]{hyperref}
\usepackage[utf8]{inputenc}
\usepackage[small]{caption}
\usepackage{graphicx}
\usepackage{amsmath}
\usepackage{amsthm}
\usepackage{booktabs}
\usepackage{algorithm}
\usepackage{algorithmic}
\usepackage[switch]{lineno}
\usepackage{pifont}
\usepackage{enumitem} 
\usepackage{amssymb}
\usepackage{booktabs} 
\usepackage{siunitx} 
\usepackage{multirow}
\usepackage{float}
\usepackage{subcaption}
\definecolor{softgreen}{RGB}{7,188,59} 
\definecolor{myyellow}{RGB}{253,190,0} 
\definecolor{mypurple}{RGB}{174,137,192} 

\begin{document}

\title{C3S3: Complementary Competition and Contrastive Selection for Semi-Supervised Medical Image Segmentation}
\author{\IEEEauthorblockN{\textit{Jiaying He$^{\dagger,1}$, Yitong Lin$^{\dagger,1}$, Jiahe Chen$^{1}$, Honghui Xu$^{\ast,1}$, Jianwei Zheng$^{\ast,1}$} 
\thanks{$^{\dagger}$Equal contributions. $^{\ast}$Corresponding author.} 
\thanks{This work was supported in part by the National Natural Science Foundation of China under Grant 62276232 and 62406285, the China Postdoctoral Science Foundation under Grant 2024M752863, and the Key Program of Natural Science Foundation of Zhejiang Province under Grant LZ24F030012.}
}
\IEEEauthorblockA{
\text{$^{1}$Zhejiang University of Technology, Hangzhou, China} \\
\text{Email:\{xhh, zjw\}@zjut.edu.cn        }\\
}
}
\maketitle
\begin{abstract}
For the immanent challenge of insufficiently annotated samples in the medical field, semi-supervised medical image segmentation (SSMIS) offers a promising solution. 
Despite achieving impressive results in delineating primary target areas, most current methodologies struggle to precisely capture the subtle details of boundaries. 
This deficiency often leads to significant diagnostic inaccuracies. To tackle this issue, we introduce C3S3, a novel semi-supervised segmentation model that synergistically integrates complementary competition and contrastive selection. This design significantly sharpens boundary delineation and enhances overall precision. Specifically, we develop an \textit{Outcome-Driven Contrastive Learning} module dedicated to refining boundary localization. Additionally, we incorporate a \textit{Dynamic Complementary Competition} module that leverages two high-performing sub-networks to generate pseudo-labels, thereby further improving segmentation quality. The proposed C3S3 undergoes rigorous validation on two publicly accessible datasets, encompassing the practices of both MRI and CT scans. The results demonstrate that our method achieves superior performance compared to previous cutting-edge competitors. Especially, on the 95HD and ASD metrics, our approach achieves a notable improvement of at least 6\%, highlighting the significant advancements. The code is available at \url{https://github.com/Y-TARL/C3S3}.
\end{abstract}


\section{Introduction}
\label{sec:intro}

Medical image segmentation (MIS) is essential in healthcare, aiding clinicians in monitoring disease progression and planning treatments. The advancement of deep learning, especially supervised methods, has greatly enhanced MIS, making it the leading approach in the field and significantly improving performance in various practices
\cite{Yang2023DirectionalCS}. 
Traditional supervised deep learning methods require a large amount of labeled data for training, and pixel-level annotation of medical images is both time-consuming and challenging. To address this issue, semi-supervised medical image segmentation techniques have been developed, aiming to reduce the need for extensive manual annotation while maintaining high performance.

\begin{figure}[htbp]
	\centering
	\begin{subfigure}{0.23\linewidth}
		\centering
		\includegraphics[width=0.9\linewidth]     {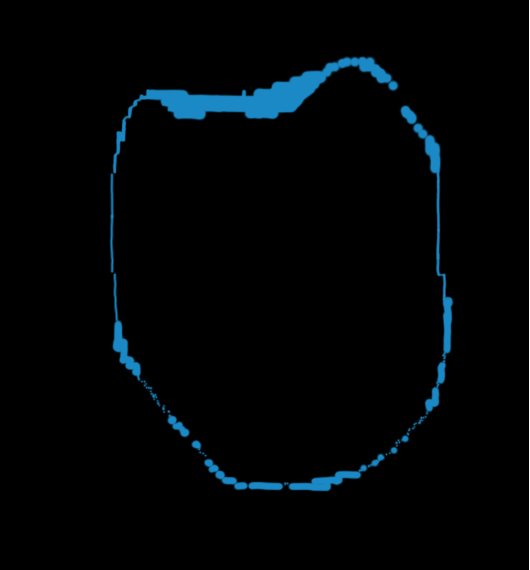}
            \scriptsize 95HD(voxel): \textbf{5.14}
            \scriptsize ASD(voxel): \textbf{1.57}
		\caption{C3S3}
		\label{C3S3-contrastive}
            
	\end{subfigure}
	\centering
	\begin{subfigure}{0.23\linewidth}
		\centering
		\includegraphics[width=0.9\linewidth]{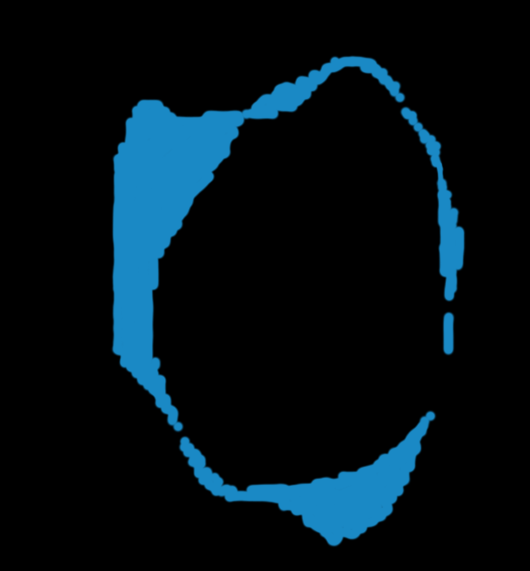}
              \scriptsize 95HD(voxel): 7.54
            \scriptsize ASD(voxel): 2.31
		\caption{UA-MT}
		\label{UAMT-contrastive}
	\end{subfigure}
         \centering
	\begin{subfigure}{0.23\linewidth}
		\centering
		\includegraphics[width=0.9\linewidth]{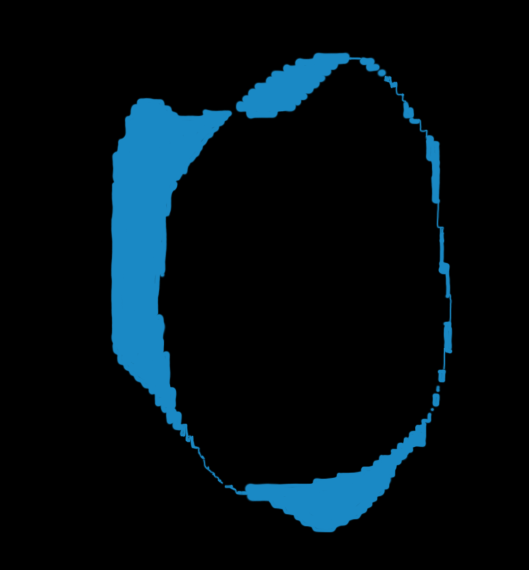}
             \scriptsize 95HD(voxel): 7.36
            \scriptsize ASD(voxel): 2.15
		\caption{DTC}
		\label{DTC-contrastive}
	\end{subfigure}
        \centering
	\begin{subfigure}{0.23\linewidth}
		\centering
		\includegraphics[width=0.9\linewidth]{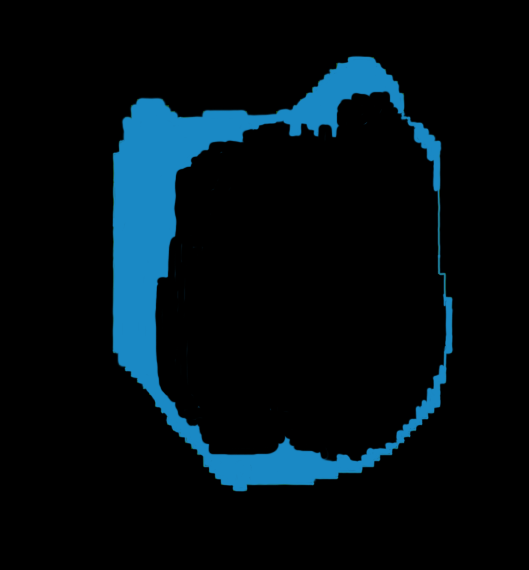}
              \scriptsize 95HD(voxel): 6.68
            \scriptsize ASD(voxel): 2.02 
		\caption{MCF}
		\label{MCF-contrastive}
	\end{subfigure}
	\caption{The visual comparison among C3S3 and others, where the blue areas represent inaccurately localized segments.}
	\label{figure1}
    \vspace{-1.2em}
\end{figure}

SSMIS primarily stands out by leveraging unlabeled data to improve segmentation performance\cite{Chen2022MagicNetSM,Bai2023BidirectionalCF}. Within this framework, two key strategies, \textit{i.e.}, pseudo-label supervision and consistency regularization, have been extensively investigated \cite{Shen2023CotrainingWH}. The former uses pseudo-labels to supervise unlabeled images, while the latter enforces prediction consistency between different models or branches, often based on popular frameworks like Mean Teacher (MT)\cite{li2020transformation,Shi2023CompetitiveET} or co-training \cite{Shen2023CotrainingWH,10688310,caussl}. These approaches typically train two networks to ensure consistent predictions, even when the same samples are exposed to random variations. Despite the success of these methods in accurately locating target regions, refining fine boundaries remains a significant challenge.

By integrating two pop techniques, \textit{i.e.}, semi-supervised learning and contrastive learning, we propose a new model to tackle the challenge. To begin with, our approach incorporates an outcome-driven contrastive learning module (ODCL), meticulously designed to enhance the delineation of boundary information. Note that contrastive learning is dedicated to train a model enjoying high discriminability of similar (positive) and dissimilar (negative) data pairs, which allows for a more nuanced understanding of the image content. On that basis, enhanced accuracy and refined details can be earned from the final model, particularly when fused with semi-supervised learning. 
Second, we elaborate a dynamic complementary competition module (DCC) to innovate beyond the traditions. Traditionally, within Mean Teacher models, the performance of the teacher network depends heavily on the student network, since the former is essentially a weighted combination of the latter's historical states. In contrast, our DCC minimizes the reliance between the student and teacher networks while utilizing pseudo-labeling to promote complementary competition, fostering a collaborative dynamic that enables mutually beneficial improvements. Figure \ref{figure1} shows an example comparison between our C3S3 and others. As can be seen, our proposal captures the much clearer boundary, with significant advantage over that from the competitors. The numerical values also verify the visual results. Technically, the primary contributions of our work are threefold.

\begin{itemize}
\item To strive for an enhanced boundary delineation, a new module named ODCL is proposed, which encompasses two key components, \textit{i.e.}, Spatial Position Binary Masking and Dual-Space Intersection-Union Loss.
\item We propose a DCC module that employs a weighted competing mechanism for precise pseudo-labeling, which also enables two subnets of the whole model to mutually learn from each other.
\item With VNet and ResVNet behaved as the dual subnets and the proposed two modules seamlessly assembled, a final segmentation network C3S3 is molded. The results on two public benchmark datasets, \textit{i.e.}, left atrium segmentation in MRI and pancreas segmentation in CT scans, verify that C3S3 sets a new state-of-the-art score.
\end{itemize}

\section{Related Work}
\subsection{Semi-supervised Medical Image Segmentation}

Recent advancements in semi-supervised medical image segmentation have introduced a variety of innovative methods. For example, CauSSL \cite{caussl} employs a novel causal graph to provide a theoretical foundation for mainstream semi-supervised segmentation approaches. Notably, two methodologies closely aligned with our research are self-training with pseudo-labeling \cite{basak2023pseudo,10688240} and consistency regularization\cite{li2020transformation,yu2019uncertainty}, both of which have demonstrated significant potential in enhancing segmentation performance.

Pseudo-labeling involves using a trained model to generate labels for unlabeled data, which are subsequently used to refine the model through iterative updates, thereby improving its learning capability. However, with the abundance of unlabeled data, the risk of error propagation caused by inaccurate pseudo-labels poses a significant challenge. Therefore, enhancing the reliability and accuracy of pseudo-labels becomes a critical priority. C3T \cite{10688240} introduces a Credibility-Aware Pseudo-Label Filtering mechanism, specifically designed for pixel-wise segmentation tasks, which validates and filters pseudo-labels to enhance inter-network guidance reliability and mitigate error propagation. Additionally, CAML \cite{caml} introduces modules that transfer prior knowledge and enforce distribution consistency, enhancing pseudo-labeling reliability.

Consistency regularization is a key technique in semi-supervised learning, enforcing consistent predictions for perturbed samples to reduce error propagation. It has been extended to multi-task learning by incorporating auxiliary tasks to utilize geometric information. In medical image segmentation (MIS), the Mean Teacher (MT) framework is widely adopted, where a student network is updated via gradients and a teacher network via EMA. TraCoCo \cite{tracoco} enhances consistency learning by varying spatial contexts, improving focus on foreground objects and segmentation performance. CORN \cite{li2024leveragingcoralcorrelationconsistencynetwork} leverages second-order statistics and a Dynamic Feature Pool to boost precision and maintain consistency. Besides, ICR-Net \cite{10688011} integrates labeled data to enhance regularization and improve information extraction from unlabeled data, complementing traditional consistency methods.

\subsection{Contrastive Learning}
Contrastive learning has shown remarkable success in the computer vision field \cite{Zhao2023RCPSRC,UGCL2024}, focusing on learning discriminative features by distinguishing similar (positive) and dissimilar (negative) feature pairs. Initially successful in image-level tasks, it has since been adapted for dense prediction tasks like image segmentation, with growing interest in semi-supervised applications. UGCL \cite{UGCL2024} improves segmentation accuracy in medical image analysis by leveraging uncertainty-aware contrastive learning to address challenges such as boundary fuzziness and high-entropy regions.

More recently, the integration of contrastive learning with consistency regularization in semi-supervised segmentation has marked a significant progress, which leverages pseudo-labels for contrastive learning while simultaneously optimizing contrastive and consistency losses \cite{Zhao2023RCPSRC}. This dual approach promotes the learning of feature representations and the refinement of pseudo-labels, thereby improving both the efficiency and accuracy of segmentation.

\section{Methodology}
\begin{figure*}[htbp]
    \centering
    \includegraphics[width=0.78\linewidth]{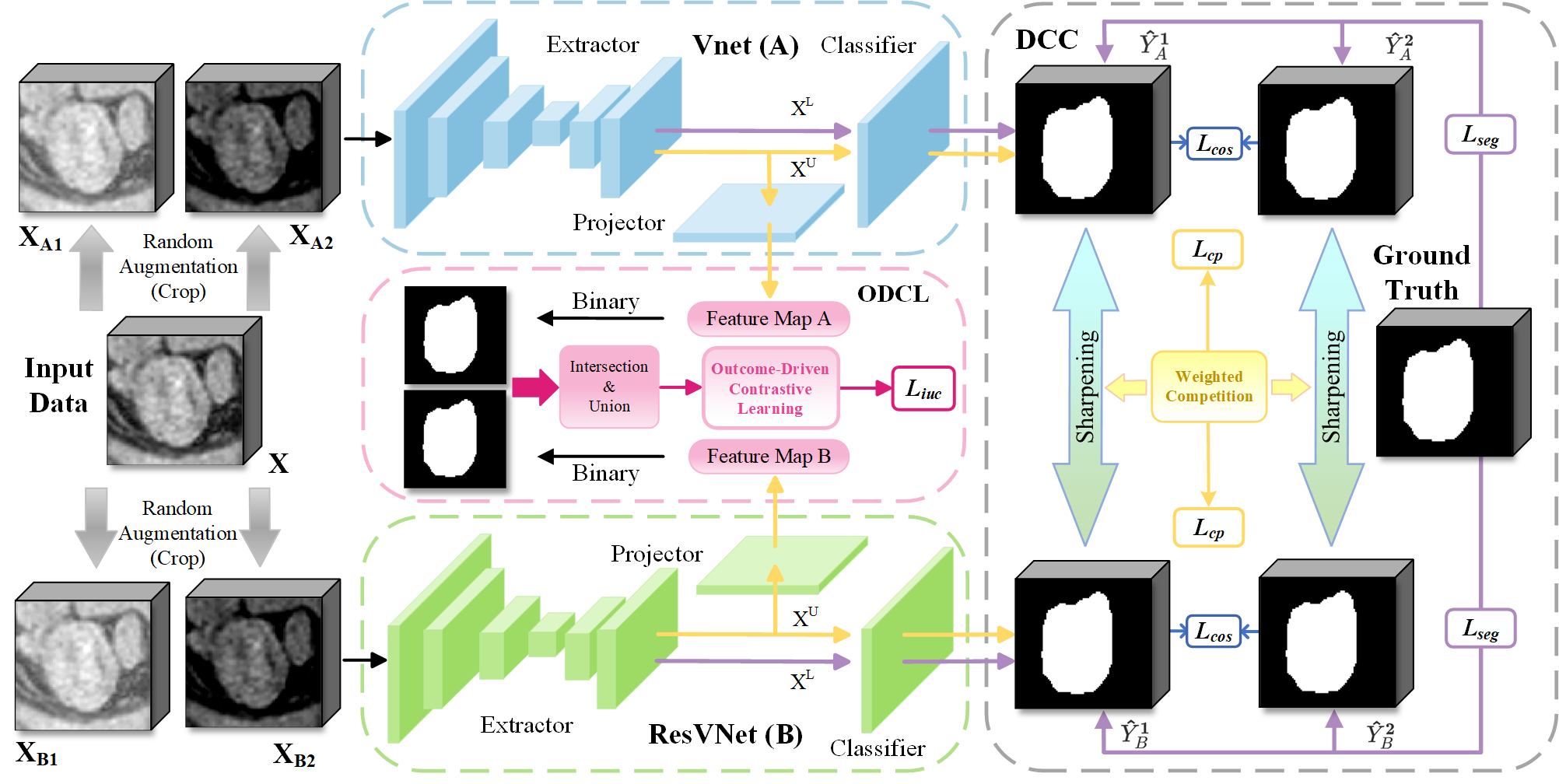}
    \caption{Overview of our proposal. Upon receiving the input data, C3S3 undergoes two random data augmentations, followed by two separate backbones, \textit{i.e.}, VNet and ResVNet. The \textcolor{myyellow}{yellow} line represents unlabeled samples, while the \textcolor{mypurple}{purple} line indicates labeled samples. The unlabeled data, after processing through the projector, is directly fed into ODCL, during which a loss $L_{iuc}$ is incurred. Both VNet and ResVNet yield two prediction results, between which a consistency loss $L_{cos}$ is computed, followed by the generation of a pseudo-label loss $L_{cp}$ via the weighted competition module. In addition, labeled data contribute to a supervised loss $L_{seg}$.}
    \label{figure2}
\end{figure*}

In accordance with MCF \cite{MCF}, the renowned VNet (A) and its variant ResVNet (B) are employed as our dual backbones, both of which own a u-shaped extractor and a tail-ready classifier. Besides, as given in Figure \ref{figure2}, the core ingredient of our proposal is centered on two principal modules, \textit{i.e.}, DCC and ODCL, whose detailed explanations are provided in Subsections \ref{sec:DCC} and \ref{sec:ODCL}, respectively.


\subsection{The overview framework}
During semi-supervised learning, it is assumed that training dataset contains \( N \) labeled data and \( M \) unlabeled data, where \( M \gg N \). For convenience, we denote a small amount of labeled data as \( \mathcal{D}_L = \{ (x_i^L, y_i^L) \}_{i=1}^N \) and the unlabeled counterpart as \( \mathcal{D}_U = \{ x_i^U \}_{i=1}^M \), where \( x_i \in \mathbb{R}^{H \times W \times D} \) denotes the training image and \( y_i \in \mathbb{B}^{H \times W \times D} \) is the label (if available). Yet with a limited number of labeled samples \( x_i^L \), the objective of semi-supervised learning is to achieve promising results with the aid of the extra unlabeled data \( x_i^U \).

With two random data augmentations initially imposed \cite{MCF}, the whole input \( \mathbf{X} \) is then individually fed into two backbones, generating the learned features for labeled data \( \hat{Y}^L \) and unlabeled data \( \hat{Y}^U \).
\begin{equation}
    \hat{Y}_A = A(\mathbf{X}), \hat{Y}_B = B(\mathbf{X})
\end{equation}

Referring to previous work \cite{Zhao2023RCPSRC}, for labeled data, cross entropy loss \( L_{ce} \) and dice loss \( L_{dice} \) are utilized for supervised training:
\begin{equation}
    L_{seg} = L_{ce}(\hat{Y}^L, y) + L_{dice}(\hat{Y}^L, y)
\end{equation} 


For unlabeled data, we calculate the pseudo supervision loss \( L_{cp} \) and outcome-driven contrastive loss \( L_{iuc} \):
\begin{equation}\label{(4)}
    L_{unsup} = L_{cp}(\hat{Y}^U_A, \hat{Y}^U_B, \hat{Y}^L) + \lambda L_{iuc}(p_A, p_B, p^-)
\end{equation}
where $\lambda$ is a weight balancing the two sub-loss, whose detailed setting is dependent on the specific task. The practices of \( p_A \), \( p_B \), and \( p^- \) will be elaborated in Subsection \ref{ODCL}. 

Meanwhile, we use consistency regularization to further reduce the difference between two predictions, leading to a consistency loss \( L_{cos} \). With all the sub-loss assembled, the overall loss is given as $L_{total} = L_{seg} + L_{cos} + L_{unsup}$.
\subsection{DCC Module} \label{sec:DCC}

We propose a dynamic complementary competition mechanism, enabling two backbones to simultaneously learn from and compete with each other. The dynamic nature of DCC allows for real-time adaptation to changing data patterns, which enjoys a significant improvement over the static version. Simultaneously, as two backbones share equal status, the decoupling of tight dependency should be accomplished, further allowing for more exclusive focus on the respective learning. 


In practice, we commence with two random augmentations of the inputs, followed by the respective feeding into backbone A and backbone B. Consequently, the concerned outputs \{\(\hat{Y}^1_A\), \(\hat{Y}^2_A\)\} and \{\(\hat{Y}^1_B\), \(\hat{Y}^2_B\)\} can be generated. After obtaining the supervised loss \(L_{seg}\) using ground truth, a weighted competition function is employed to determine which backbone performs more competitively in the current state. 
\begin{equation}
    L_{competition} = \alpha L_{ce}(\hat{Y}^L, y) + (1 - \alpha) L_{dice}(\hat{Y}^L, y)
\end{equation}
where parameter \(\alpha\) is the determining weight. 

Then, we utilize the currently better-performing backbone to generate pseudo-labels, supervising the predictions of the other backbone and computing the pseudo supervised loss:
\begin{equation}
    L_{cp} = MSE(\hat{Y}^{U1}, Y^{U1}_p) + MSE(\hat{Y}^{U2}, Y^{U2}_p)
\end{equation}

Meanwhile, to minimize the disagreement between \(\hat{Y}^1_A\) and \(\hat{Y}^2_A\), (Resp. \(\hat{Y}^1_B\) and \(\hat{Y}^2_B\)), the cosine distance between two predicted results is utilized to obtain a consistency loss:
\begin{equation}
    L_{cos}(\hat{Y}^1_A, \hat{Y}^2_A) = 1 - \frac{\hat{Y}^1_A \cdot \hat{Y}^2_A}{\|\hat{Y}^1_A\|_2 \cdot \|\hat{Y}^2_A\|_2}
\end{equation}


The integration of an explicit loss term into the model offers additional regularization during training, which is vital to improving the stability and consistency of the entire training process. Additionally, during the selection of the better-performed backbone, since the two loss terms in the competition are already calculated during the model training, hence introducing additional computations is not required, which further guarantees a reduced computational overhead.

\subsection{ODCL Module} \label{ODCL}   \label{sec:ODCL}

\begin{figure}[htbp]
    \centering
    \includegraphics[width=0.9\linewidth]{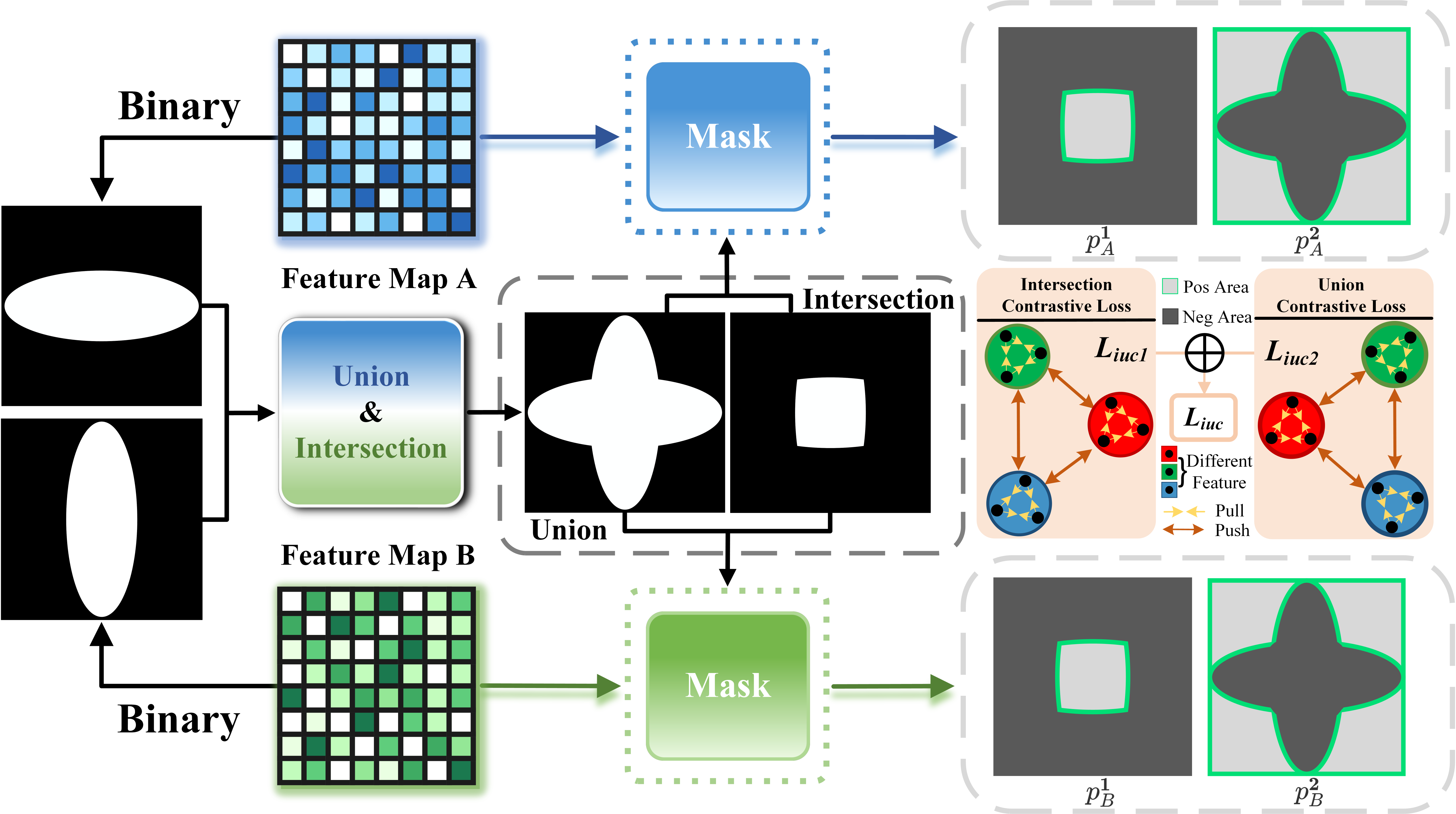}
    \caption{Illustration of ODCL.
    ODCL identifies positive and negative sample regions from feature maps processed by intersection and union operations, then calculates the loss $L_{iuc}$.}
    \label{figure3}
\end{figure}

To ensure consistency between the feature spaces of two backbones, contrastive learning is integrated, which operates in two phases: spatial position binary masking and dual-space intersection-union loss. Contrastive learning enables the model to gain a deeper understanding of complex data structures, offering more detailed data analysis.

\textbf{Spatial Position Binary Masking:} Figure \ref{figure3} provides an overview of the mask sampling scheme. Unlabeled data is firstly input into two backbones, followed by the projection of output from the extractor into low-dimensional feature maps. Subsequently, we binarize the two feature maps, with the intersection and union of the results calculated to derive the masks. The masks from the intersection and the union are respectively labeled as \(\mathcal{M}_{\cap}\) and \(\mathcal{M}_{\cup}\).
With the feature maps passed through the two masks, the results obtained are denoted as \( p_A^1 \) and \( p_A^2 \) (Resp. \( p_B^1 \) and \( p_B^2 \)), with negative samples defined as \( p^-\). Note that these notation has appeared in Eq. \ref{(4)}.

We consider the intersection part as the correct foreground prediction by a high probability, and the voxels excluding the union as the correct background prediction. When we set positive samples from the foreground region, the negative sample would be selected from the possible background region, and vise versa, which ensures an evident contrasting effect.

In practice, we design a strategy for selecting positive and negative sample pairs. That is, given an anchor, the corresponding voxel at the same spatial position is chosen as the positive sample. With the utilization of the intersection mask, we further designate the intersecting area as positive sample area and the remaining as negative samples, from which the contrastive loss \(L_{iuc1}\) can be calculated. Similarly, with the union mask, we consider the voxels excluding the union result as positive sample area and the rest as negative samples to compute the contrastive loss \(L_{iuc2}\).

\textbf{Dual-Space Intersection-Union Loss:} Following the respective positive and negative sampling, a contrastive loss is formulated to draw positive pairs closer and negative pairs far away. Specifically, for a foreground voxel feature \(\varphi_1 \in p_A\), the contrastive loss is computed as follows:
\vspace{-0.2em}
\begin{equation}
  \begin{aligned}
    & L_{iuc1}(\varphi_1, \varphi_2) = -\log \\
    & \left( \frac{\mathcal{M}_{\cap} e^{\cos(\varphi_1, \varphi_2)/\tau}}{\mathcal{M}_{\cap} e^{\cos(\varphi_1, \varphi_2)/\tau} + (1 - \mathcal{M}_{\cap}) \sum_{n=1}^{N} e^{\cos(\varphi_1, \varphi_n)/\tau}} \right)
  \end{aligned}
\end{equation}
\vspace{-0.2em}
where \(\varphi_2\) represents the positive voxel extracted from \(p_B\), while \(\varphi_n\) refers to the negative voxels. \(N\) indicates the total count of sampled negative voxels, and \(\tau\) stands for the temperature hyperparameter.

Similarly, for the background voxels, the contrastive loss can be formulated as follows:
\vspace{-0.2em}
\begin{equation}
    \begin{aligned}
        & L_{iuc2}(\varphi_1, \varphi_2) = -\log \\
        & \left( \frac{(1 - \mathcal{M}_{\cup}) e^{\cos(\varphi_1, \varphi_2)/\tau}}{(1 - \mathcal{M}_{\cup}) e^{\cos(\varphi_1, \varphi_2)/\tau} + \mathcal{M}_{\cup} \sum_{n=1}^{N} e^{\cos(\varphi_1, \varphi_n)/\tau}} \right)
    \end{aligned}
\end{equation}


Therefore, the whole contrastive loss between two feature maps can be given as:
\vspace{-0.2em}
\begin{equation}
    \begin{aligned}
    & L_{iuc}(p_A, p_B, p^{-}) = \\
    & \frac{1}{N_p} \sum_{\varphi_1 \in p_A} L_{iuc1}(\varphi_1, \varphi_2)
     \quad + \frac{1}{N_p} \sum_{\varphi_1 \in p_A} L_{iuc2}(\varphi_1, \varphi_2)
    \end{aligned}
\end{equation}
where \( N_p \) denotes the number of voxels in \( p_A \).

With four feature maps from two backbones, we maximize their information utility by computing contrastive loss between feature map pairs from different branches, culminating in the definition of the dual-space intersection-union loss:
\begin{equation}
    \begin{aligned}
    & L_{all}(p_A,p_B,p^-) =  \ L_{iuc}(p_A^1,p_B^1,p^-) + L_{iuc}(p_A^1,p_B^2,p^-) \\
    & + L_{iuc}(p_A^2,p_B^1,p^-) + L_{iuc}(p_A^2,p_B^2,p^-)
    \end{aligned}
\end{equation}

\section{Experiments}

\begin{table*}[htbp] 
\centering 
\caption{Performance comparison on LA Dataset. The best and second-best performers are highlighted by bold and underline, respectively.} 
\label{table1} 
\begin{tabular}{c|cc|cccc}
\hline \multirow{2}{*}{ \textbf{Method} } & \multicolumn{2}{c|}{\textbf{Volumes used}} & \multicolumn{4}{c}{\textbf{Metrics}} \\  \cline{2-7}
& Labeled & Unlabeled & $95HD$(voxel)$\downarrow$ & $ASD$(voxel)$\downarrow$ & $Dice$(\%)$\uparrow$ & $Jaccard$(\%)$\uparrow$  \\
\hline VNet & $80(100\%)$ & $0$  & $5.61_{ }$ & $1.51_{ }$ &$91.47_{ }$& $84.36_{ }$ \\
 ResVNet & $80(100\%)$ & $0$  & $4.97_{ }$ & $1.81_{ }$ &$91.05_{ }$& $83.48_{ }$ \\
VNet & $16(20\%)$ & $0$  & $14.82_{ }$ & $3.91_{ }$ &$86.33_{ }$& $73.56_{ }$ \\
ResVNet & $16(20\%)$ & $0$  & $17.38_{ }$ & $5.02_{ }$ &$85.61_{ }$& $73.09_{ }$ \\
\hline UA-MT (MICCAI 2019) & $16(20\%)$ & $64$  & $7.54_{ }$ & $2.31_{ }$ &$88.82_{ }$& $80.18_{ }$ \\
SASSNet (MICCAI 2020) & $16(20\%)$ & $64$  & $8.75_{ }$ & $3.10_{ }$ &$89.29_{ }$& $80.85_{ }$ \\
DTC (AAAI 2021) & $16(20\%)$ & $64$  & $7.36_{ }$ & $2.15_{ }$ &$89.45_{ }$& $81.01_{ }$ \\
SS-Net (MICCAI 2022) & $16(20\%)$ & $64$  & $6.96_{ }$ & $1.84_{ }$ &$90.21_{ }$& $82.36_{ }$ \\
MCF (CVPR 2023) & $16(20\%)$ & $64$  & $6.68_{ }$ & $2.02_{ }$ &$88.02_{ }$& $79.51_{ }$ \\
CAML (MICCAI 2023) & $16(20\%)$ & $64$  & $6.11_{ }$ & $\underline{1.68}_{ }$ &$90.78_{ }$& $83.19_{ }$ \\ 
TraCoCo(TMI 2024)& $16(20\%)$ & $64$  & $\underline{5.63}$ & $1.79$ &$\underline{90.94}_{ }$& $\underline{83.47}_{ }$ \\ 
\textbf{C3S3 (ours)}  & $16(20\%)$ & $64$ & $\textbf{5.14}$ & $\textbf{1.57}$ & $\textbf{91.24}$ & $\textbf{84.01}$ \\
\hline
\end{tabular}
\end{table*}
We evaluate our approach on two datasets: the Left Atrial (LA) dataset \cite{xiong2020global} with 100 MR volumes (\(112 \times 112 \times 80\) crops for training and sliding windows for inference) and the Pancreas-CT dataset \cite{Roth2015DeepOrganMD} with 82 CT scans (\(96 \times 96 \times 96\) crops and $K$-fold cross-validation for evaluation).

In line with recent state-of-the-art methods \cite{Bai2023BidirectionalCF,MCF}, we employ four widely used metrics: Dice Similarity Coefficient (Dice), Jaccard Similarity Coefficient (Jaccard), 95\% Hausdorff Distance (95HD), and Average Surface Distance (ASD). Dice and Jaccard evaluate regional overlap between predicted segmentations and ground truth, while 95HD and ASD assess boundary alignment. Specifically, 95HD measures the 95th percentile of surface-to-surface distances, and ASD computes the average surface distance. For comparison, we benchmark against several advanced methods, including UA-MT \cite{yu2018pu}, SASSNet \cite{Li2020ShapeawareS3}, DTC \cite{luo2021semi}, SS-Net \cite{wu2022exploring}, MCF \cite{MCF}, CAML \cite{caml}, CauSSL \cite{caussl}, and TraCoCo \cite{tracoco}.

\begin{table*}[htbp] 
\centering 
\caption{Comparisons on Pancreas-CT Dataset. The best and second-best performers are highlighted by bold and underline, respectively.} 
\label{table2} 
\begin{tabular}{c|cc|cccc}
\hline \multirow{2}{*}{ \textbf{Method} } & \multicolumn{2}{c|}{\textbf{Volumes used}} & \multicolumn{4}{c}{\textbf{Metrics}} \\  \cline{2-7}
& Labeled & Unlabeled & $95HD$(voxel)$\downarrow$ & $ASD$(voxel)$\downarrow$ & $Dice$(\%)$\uparrow$ & $Jaccard$(\%)$\uparrow$  \\

\hline VNet & $62(100\%)$ & $0$  & $7.21_{ }$ & $1.69{ }$ &$80.72_{ }$& $68.90_{ }$ \\
 ResVNet & $62(100\%)$ & $0$  & $7.38_{ }$ & $1.63_{ }$ &$79.84_{ }$& $67.35_{ }$ \\
VNet & $12(20\%)$ & $0$  & $17.69_{ }$ & $4.64_{ }$ &$64.04_{ }$& $48.17_{ }$ \\
ResVNet & $12(20\%)$ & $0$  & $18.95_{ }$ & $5.63_{ }$ &$66.45_{ }$& $51.12_{ }$ \\
\hline UA-MT (MICCAI 2019) & $12(20\%)$ & $50$  & $17.06_{ }$ & $5.23_{ }$ &$74.02_{ }$& $60.10_{ }$ \\
SASSNet (MICCAI 2020) & $12(20\%)$ & $50$  & $13.89_{ }$ & $3.56_{ }$ &$73.58_{ }$& $59.73_{ }$ \\
DTC (AAAI 2021) & $12(20\%)$ & $50$  & $13.18_{ }$ & $3.75_{ }$ &$73.31_{ }$& $59.26_{ }$ \\
SS-Net (MICCAI 2022) & $12(20\%)$ & $50$  & $12.64_{ }$ & $3.59_{ }$ &$74.27_{ }$& $60.56_{ }$ \\
MCF (CVPR 2023) & $12(20\%)$ & $50$  & $11.59_{ }$ & $3.27_{ }$ &$75.00_{ }$& $61.27_{ }$ \\
CauSSL (ICCV 2023) & $12(20\%)$ & $50$  & $8.11_{ }$ & $2.34_{ }$ &$79.97_{ }$& $67.32_{ }$ \\
TraCoCo(TMI 2024)& $12(20\%)$ & $50$  & $\underline{7.34}_{ }$ & $\underline{1.84}_{ }$ &$\underline{80.90}_{ }$& $\textbf{69.26}$ \\ 
\textbf{C3S3 (ours)}  & $12(20\%)$ & $50$ & $\textbf{6.96}$ & $\textbf{1.74}$ & $\textbf{80.93}$ & \underline{$68.06$} \\
\hline
\end{tabular}
\end{table*}

\begin{figure*}[htbp]
    \centering
    \begin{subfigure}{\textwidth}
        \centering
        \begin{subfigure}{0.12\linewidth}
            \centering
            \includegraphics[width=0.9\linewidth]{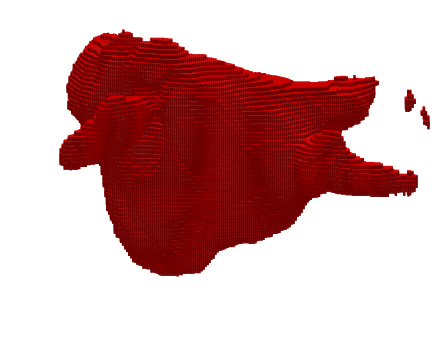}
           \small UA-MT
        \end{subfigure}
        \begin{subfigure}{0.12\linewidth}
            \centering
            \includegraphics[width=0.9\linewidth]{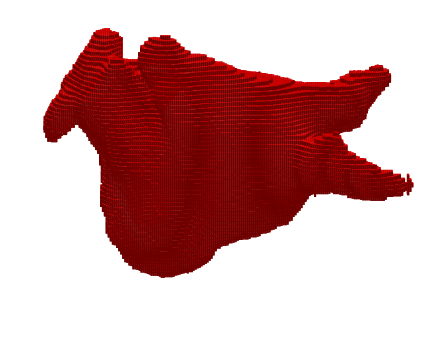}
           \small DTC
        \end{subfigure}
        \begin{subfigure}{0.12\linewidth}
            \centering
            \includegraphics[width=0.9\linewidth]{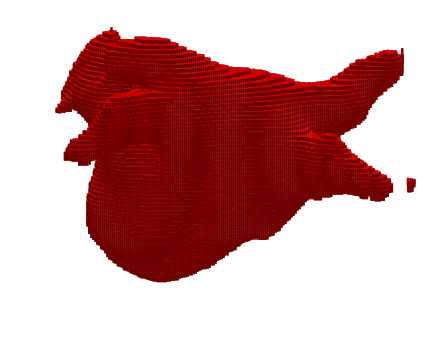}
           \small MCF
        \end{subfigure}
        \begin{subfigure}{0.12\linewidth}
            \centering
            \includegraphics[width=0.9\linewidth]{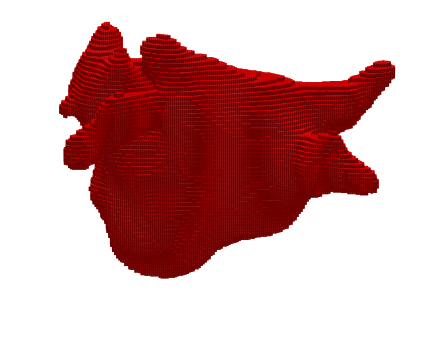}
           \small CAML
        \end{subfigure}
        \begin{subfigure}{0.12\linewidth}
            \centering
            \includegraphics[width=0.9\linewidth]{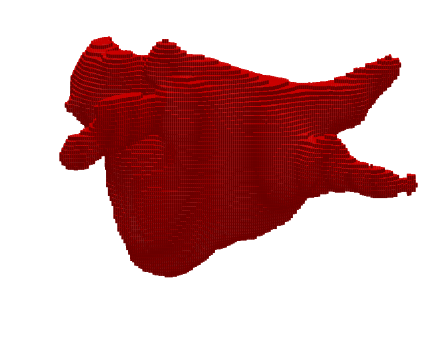}
           \small TraCOCO
        \end{subfigure}
        \begin{subfigure}{0.12\linewidth}
            \centering
            \includegraphics[width=0.9\linewidth]{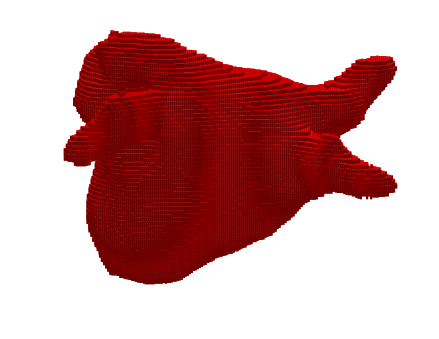}
           \small C3S3
        \end{subfigure}
        \begin{subfigure}{0.12\linewidth}
            \centering
            \includegraphics[width=0.9\linewidth]{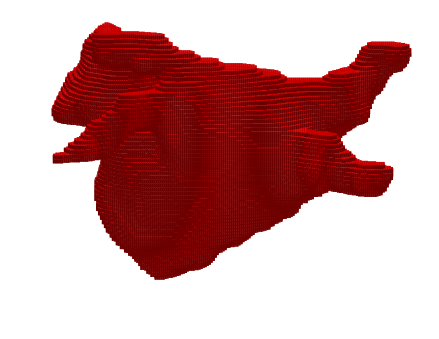}
           \small Ground Truth
        \end{subfigure}
        \caption{Results on the LA Dataset.}
    \end{subfigure}


    \begin{subfigure}{\textwidth}
        \centering
        \begin{subfigure}{0.12\linewidth}
            \centering
            \includegraphics[width=0.9\linewidth]{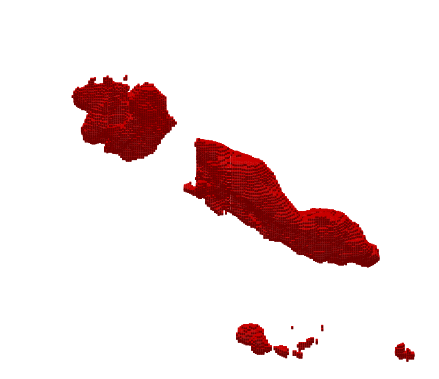}
          \small  UA-MT
        \end{subfigure}
        \begin{subfigure}{0.12\linewidth}
            \centering
            \includegraphics[width=0.9\linewidth]{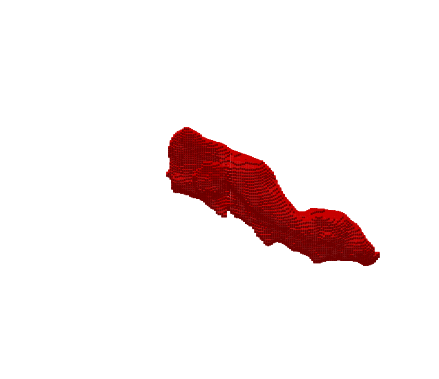}
          \small  DTC
        \end{subfigure}
        \begin{subfigure}{0.12\linewidth}
            \centering
            \includegraphics[width=0.9\linewidth]{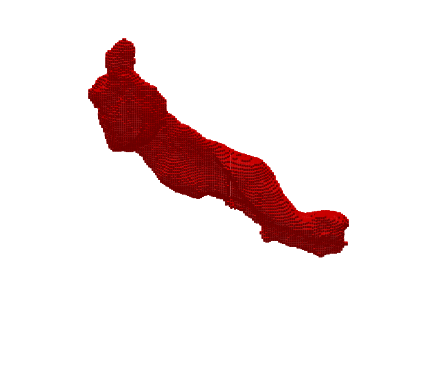}
           \small MCF
        \end{subfigure}
        \begin{subfigure}{0.12\linewidth}
            \centering
            \includegraphics[width=0.9\linewidth]{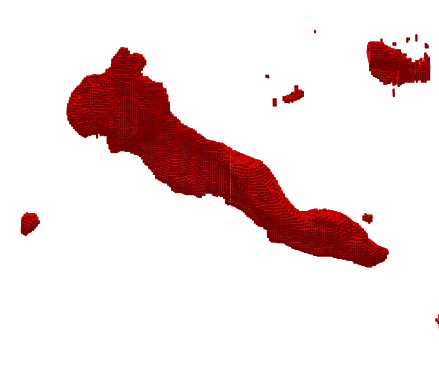}
           \small CauSSL
        \end{subfigure}
        \begin{subfigure}{0.12\linewidth}
            \centering
            \includegraphics[width=0.9\linewidth]{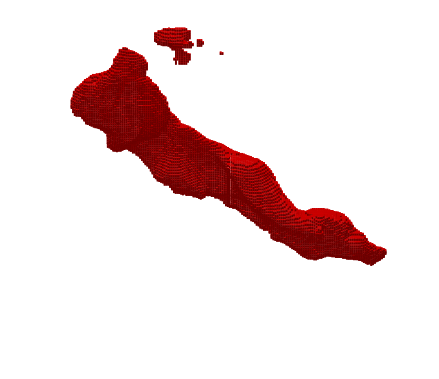}
          \small TraCOCO
        \end{subfigure}
        \begin{subfigure}{0.12\linewidth}
            \centering
            \includegraphics[width=0.9\linewidth]{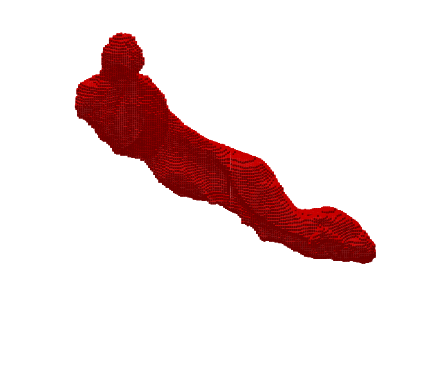}
           \small C3S3
        \end{subfigure}
        \begin{subfigure}{0.12\linewidth}
            \centering
            \includegraphics[width=0.9\linewidth]{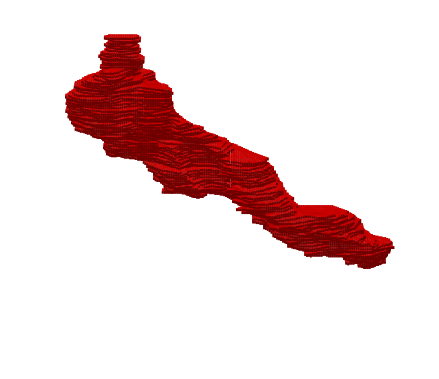}
           \small Ground Truth
        \end{subfigure}
        \caption{Results on the Pancreas-CT Dataset.}
    \end{subfigure}
\vspace{-1em}
    \caption{Qualitative segmentation results. (a) Results on the LA dataset. (b) Results on the Pancreas-CT dataset.}
    \label{figure4}
    \vspace{-1.5em}
\end{figure*}
\vspace{-0.1em}
\subsection{Results on LA}
\vspace{-0.1em}

Table \ref{table1} presents the evaluation results on the LA dataset, including comparisons with standalone VNet and ResVNet under fully supervised (100\% labeled) and semi-supervised (20\% labeled) settings. C3S3 achieves a 95HD of 5.14, outperforming the fully supervised ResVNet (5.61) and closely approaching VNet (4.97). For ASD, C3S3 achieves 1.57, the closest to fully supervised performance. Among semi-supervised methods, C3S3 demonstrates substantial improvements over competitors. Notably, C3S3 achieves the highest Dice (91.24\%), surpassing semi-supervised ResVNet (91.05\%) and nearly matching fully supervised VNet (91.47\%). Similarly, its Jaccard index (84.01\%) is comparable to VNet (84.36\%). These results highlight the precise boundary localization and segmentation accuracy of C3S3. Figure \ref{figure4} visually confirms these findings, demonstrating that C3S3 captures finer details with superior boundary delineation.

\vspace{-0.1em}
\subsection{Results on Pancreas-CT}
\vspace{-0.1em}

Table \ref{table2} presents the evaluation results on the Pancreas-CT dataset, a more challenging task than LA MRIs due to the complex background of pancreatic CT volumes. The performance of C3S3 is benchmarked against VNet and ResVNet under both fully supervised (100\% labeled data) and semi-supervised (20\% labeled data) settings. A four-fold cross-validation strategy is applied, using 12 labeled and 50 unlabeled samples to ensure a robust evaluation. C3S3 achieves outstanding results, particularly in boundary-sensitive metrics, with 95HD reduced to 6.96 and ASD to 1.74, significantly outperforming VNet and ResVNet trained with 20\% labeled data. Compared to other semi-supervised models, C3S3 also achieves the best results in 95HD and ASD, highlighting its superior boundary delineation capabilities. In terms of Dice and Jaccard metrics, C3S3 achieves the highest Dice score (80.93\%) and the second-highest Jaccard index (68.06\%) among all semi-supervised methods. Figure \ref{figure4} visually illustrates these results, showcasing C3S3's precise boundary localization and improved segmentation accuracy compared to other approaches.

\section{Ablation Study}
\textbf{Influence of hyperparameter $\alpha$:} In the DCC module, a weighted competition evaluates the practical performance of two backbone networks, selecting the most competitive one for pseudo-labeling. Analysis in the left atrial dataset shows that different values of the hyperparameter $\alpha$ have a relatively mild impact on model performance, with the best results achieved when $\alpha$ is set to 0.8, as shown in Table \ref{table3}.

\begin{table}[htbp]
\centering 
\caption{Influence of different $\alpha$ values in the weighted competition mechanism on LA Dataset.
} 
\label{table3} 
\resizebox{0.48\textwidth}{!}{ 
\begin{tabular}{c|cccc}
\hline 
\multicolumn{1}{c|}{\textbf{Parameter}} & \multicolumn{4}{c}{\textbf{Metrics}} \\
\hline 
$\alpha$ &  $95HD$(voxel)$\downarrow$ & $ASD$(voxel)$\downarrow$ & $Dice$(\%)$\uparrow$ & $Jaccard$(\%)$\uparrow$ \\
\hline 
$1.0$                        &5.94      &1.97    &88.89     &80.58         \\
$0.9$                      &5.62      &1.90    &89.70     &81.57         \\
$0.8$                      &\textbf{5.14}      &\textbf{1.57}    &\textbf{91.24}     &\textbf{84.01}         \\
$0.7$                      &5.40      &1.66   &90.34     &82.56         \\
$0.6$                      &5.87      &1.84    &89.17     &80.91         \\
$0.5$                      &5.83      &1.88    &89.59     &81.44         \\
$0.4$                      &5.76      &2.00    &89.26     &80.97         \\
$0.3$                      &7.14      &2.16    &89.15     &80.78         \\
$0.2$                      &5.83      &2.02    &88.51     &80.20        \\
$0.1$                      &5.75      &1.95    &89.91     &81.85         \\
$0.0$                      &6.47      &2.01    &88.70     &80.27         \\
\hline
\end{tabular}
} 
\end{table}

\begin{table}[htbp]
\centering 
\caption{Ablation results of ODCL and DCC Module.
} 
\label{table4} 
\resizebox{0.48\textwidth}{!}{ 
\begin{tabular}{c|cc|cccc} 
\hline 
\multirow{2}{*}{\textbf{Method}} & \multirow{2}{*}{$L_{iuc}$} & \multirow{2}{*}{$L_{cp}$} & \multicolumn{4}{c}{\textbf{Metrics}} \\
                & &  & $95HD$(voxel)$\downarrow$ & $ASD$(voxel)$\downarrow$ & $Dice$(\%)$\uparrow$ & $Jaccard$(\%)$\uparrow$ \\
\hline 
(1)                   &$-$             &$-$             &6.97     &2.03     &87.12      &80.14         \\
(2)                   &$\checkmark$    &$-$             &5.64     &1.81     &89.92      &81.88         \\
(3)                   &$-$             &$\checkmark$    &6.27      &1.86     &90.18      &82.26        \\

(4)                   &$\checkmark$    &$\checkmark$    &\textbf{5.14 }     &\textbf{1.57}     & \textbf{91.24}     & \textbf{84.01}         \\
\hline
\end{tabular}
} 
\vspace{-1em}
\end{table}



\textbf{The Role of ODCL:} The ODCL module aims to enhance precision in target boundary delineation. Experimental comparisons between models with and without ODCL demonstrate significant difference, as shown in Table \ref{table4}. Specifically, the 95HD decreases from 6.97 to 5.64, ASD drops from 2.03 to 1.81, the Dice coefficient improves from 87.12\% to 89.92\%, and the Jaccard index increases from 80.14\% to 81.88\%. These results confirm that ODCL effectively improves the accuracy of target localization across various metrics.


\textbf{The Role of DCC:} A comprehensive ablation analysis highlights the impact of the DCC module in the C3S3 framework. DCC refines predictions by dynamically generating pseudo-labels through a competitive mechanism between dual backbones, using the better-performing backbone to supervise the other. This approach improves boundary delineation accuracy and enhances both boundary-sensitive and overlap metrics. As shown in Table \ref{table4}, for example, 95HD decreases from 6.97 to 6.27, ASD reduces from 2.03 to 1.86, while Dice and Jaccard scores increase to 90.18\% and 82.26\%, respectively. These findings demonstrate DCC’s critical role in reducing pseudo-labeling errors and enhancing framework robustness. Additionally, its integration with other loss components further enhances the stability and robustness of the overall framework.
\section{Conclusion}
In this study, we propose a new semi-supervised method for 3D medical image segmentation, which is specifically designed to provide refined boundaries and then support enriched details. For that purpose, an outcome-driven contrastive learning module is meticulously developed, which further encompasses a spatial position binary masking scheme and a dual-space intersection-union loss, both of which significantly improve the boundary molding. Additionally, we have implemented a dynamic complementary competition module, which generates pseudo-labels by selecting the more competitive backbone via a weighted competition mechanism. Consistency regularization is also employed to reduce discrepancies between different predictions. The effectiveness of our C3S3 in SSMIS is validated on two public benchmark datasets. Concretely, new state-of-the-art scores have been achieved in terms of the vast majority of metrics. Moreover, we find that the proposed ODCL and DCC modules can behave as a plug-and-play role and be applicable across various fields.
\bibliographystyle{IEEEbib}
\bibliography{icme2025references}


\end{document}